\documentclass[twocolumn,10pt]{asme2ej}

\usepackage{graphicx}
\usepackage[cmex10]{amsmath}
\usepackage{array}
\usepackage{subfig}
\usepackage{epstopdf}
\usepackage{url}
\usepackage{float}
\usepackage{commath}
\usepackage{algorithm}
\usepackage{algpseudocode}
\usepackage{breqn}
\usepackage{color}
\newtheorem{theorem}{Theorem}[section]
\newtheorem{property}[theorem]{Property}

\title{Quadrotor Manipulation System: Development of a Robust Contact Force Estimation and Impedance Control Scheme Based on DOb and FTRLS}

\author{Ahmed Khalifa
    \affiliation{Cardiff School of technologies, Cardiff\\ Metropolitan University, Cardiff, UK.\\	
    	On leave from Department of Industrial\\ Electronics and Control Engineering,\\
    	Faculty of Electronic Engineering,\\ Menoufia University, Al Menoufia,  Egypt.\\
    }	
}

\author{Mohamed Fanni \\
    \affiliation{Was with Department of Mechatronics\\ and Robotics Engineering\\
	Egypt-Japan University\\ of Science and Technology\\
	Alexandria, Egypt \\
    }
}

\author{Alaa Khalifa
	\affiliation{
		Department of Industrial Electronics and Control Engineering,\\
		Faculty of Electronic Engineering,
		Menoufia University, Al Menoufia,  Egypt.\\
		Email: alaa.khalifa@el-eng.menofia.edu.eg
	}	
}

\begin{document}

\maketitle    
\sloppy
\begin{abstract}
{\it The research on aerial manipulation systems has been increased rapidly in recent years. These systems are very attractive for a wide range of applications due to their unique features. However, dynamics, control and manipulation tasks of such systems are quite challenging because they are naturally unstable, have very fast dynamics, have strong nonlinearities, are very susceptible to parameters variations due to carrying a payload besides the external disturbances, and have complex inverse kinematics. In addition, the manipulation tasks require estimating (applying) a certain force of (at) the end-effector as well as the accurate positioning of it. Thus, in this article, a robust force estimation and impedance control scheme is proposed to address these issues. The robustness is achieved based on the Disturbance Observer (DOb) technique. Then, a tracking and performance low computational linear controller is used. For teleoperation purpose, the contact force needs to be identified. However, the current developed techniques for force estimation have limitations because they are based on ignoring some dynamics and/or requiring of an indicator of the environment contact. Unlike these techniques, we propose a technique based on linearization capabilities of DOb and a Fast Tracking Recursive Least Squares (FTRLS) algorithm. The complex inverse kinematics problem of such a system is solved by a Jacobin based algorithm. The stability analysis of the proposed scheme is presented. The algorithm is tested to achieve tracking of task space reference trajectories besides the impedance control. The efficiency of the proposed technique is enlightened via numerical simulation.
}
\end{abstract}

\section{Introduction}

Recently, Unmanned Aerial Vehicles (UAVs) especially multi-rotors type, receive great attention due to their higher degree of mobility, speed and capability to access to regions that are inaccessible to ground vehicles. However, UAV as a standalone vehicle has a limited functionality to the search and surveillance applications. 

Due to their superior mobility, much interest is given to utilize them for aerial manipulation and thus the application of UAV manipulation systems have been expanded dramatically. Applications of such systems include inspection, maintenance, structure assembly,firefighting, rescue operation, surveillance, or transportation in locations that are inaccessible, very dangerous or costly to be accessed from the ground.

Research on quadrotor-based aerial manipulation can be divided into different approaches based on the tool attached to the UAV including gripper based \cite{mellinger2011design}, cables based \cite{goodarzi2015geometric, guerrero2017swing}, multi-DoF robotic manipulator based\cite{kim2013aerial, fanni2017new}, multi-DoF dual-arms manipulator based \cite{korpela2014towards}, compliant manipulator -based \cite{bartelds2016compliant}, Hybrid rigid/elastic-joint manipulator \cite{yuksel2016aerial}.

In the gripper/ tool-based approach, the attitude of the payload/tool is restricted to that of the quadrotor, and hence, the resulting aerial system has independent 4 DOFs; three translational DOFs and one rotational DOF (Yaw), i.e., the gripper/tool cannot posses pitch or roll rotation without moving horizontally. The second approach is to suspend a payload with cables but this approach has a drawback that the movement of the payload cannot be always regulated directly.  
To cope up with these limitations, another approach is developed in which a quadrotor is equipped with a robotic manipulator that can actively interact with the environment. Very few reports exist in the literature that investigate the combination of aerial vehicle with robotic manipulator. Kinematic and dynamic models of the quadrotor combined with arbitrary multi-DOF robot arm are derived using the Euler-Lagrangian formalism in \cite{lippiello2012cartesian}. In \cite{orsag2013modeling}, a quadrotor with light-weight manipulators, three 2-DOF arms, are tested. In \cite{kim2013aerial}, an aerial manipulation using a quadrotor with a 2-DOF robotic arm is presented but with certain topology that disable the system from making arbitrary position and orientation of the end-effector. In this system, the axes of the manipulator joints are parallel to each other and parallel to one in-plane axis of the quadrotor. Thus, the system cannot achieve orientation around the second in-plane axis of the quadrotor without moving horizontally. 

From the above discussion, the current introduced systems in the literature that use a gripper suffers from the limited allowable DOFs of the end-effector. The other systems have a manipulator with either two DOFs but in certain topology that disables the end-effector to track arbitrary 6-DOF trajectory, or more than two DOFs which decreases greatly the possible payload carried by the system. 

In \cite{new_quad_manp, khalifa2013adaptive, fanni2017new}, the authors propose a new aerial manipulation system that consists of 2-link manipulator, with two revolute joints whose axes are perpendicular to each other and the axis of the first joint is parallel to one in-plane axis of the quadrotor. Thus, the end-effector is able to reach arbitrary position and orientation without moving horizontally with minimum possible actuators.  

In order to achieve position holding during manipulation, uncertainties and disturbances in the system such as wind, contact forces, measurement noise have to be compensated by using a robust control scheme. Disturbance Observer (DOb)-based controller is used to achieve a robust motion control \cite{li2014disturbance, chen2012disturbance}. The DOb estimates the nonlinear terms  and uncertainties then compensates them such that the robotic system acts like a multi-SISO linear systems. Therefore, it is possible to rely on a standard linear controller to design the controller of the outer loop such that the system performance can be adjusted to achieve desired tracking accuracy and speed. In \cite{sariyildiz2015nonlinear, choi2014simplified, dong2014high}, DOb-based motion control technique is applied to robotic-based systems and gives efficient results.

In the motion control of the aerial manipulator, achievement of the compliance control is very important because the compliance motion makes possible to perform flexible motion of the manipulator according to desired impedance \cite{barbalata2018position}. This is very critical demand in applications such as demining and maintenance. In  the  compliance control, end-effector  position  and  generated  force of  the manipulator  are controlled  according to the reaction  force detected  by  the force sensor. In this method, the desired impedance is selected arbitrary in the controller. However, the force sensor is essential to detect the reaction force as presented in \cite{seraji1997force, love1995environment, singh1995analysis}. On line identified environment impedance has also been used for transparency in teleoperation systems \cite{misra2006environment}. These problems are more severe when environment displays sudden changes in its dynamic parameters which cannot be tracked by the identification process. In \cite{hashtrudi1996adaptive}, it is found that in order to faithfully convey to the operator the sense of high frequency chattering of contact between the slave and hard objects, faster identification and structurally modified methods were required. However, these methods need the measurement of force. 

Several techniques are proposed to estimate the contact force and the environment dynamics. In \cite{sariyildiz2014guide}, the DOb and Recursive Least Squares (RLS) are used to estimate the environment dynamics. However, in this method, two DObs are used besides the RLS, and the estimation of contact force is activated only during the instance of contacting, thus there is a need to detect the instant at which the contact occurs. However, this  is not practical approach especially if we target autonomous system. In \cite{murakami1993force, eom1998disturbance, van2011estimating, phong2012external, colome2013external, alcocera2004force}, several techniques are proposed to achieve force control without measuring the force. However, these techniques are based on ignoring some dynamics and external disturbances which will produce inaccurate force estimation. In \cite{forte2012impedance, lippiello2012exploiting}, an impedance control is designed for aerial manipulator without the need to measure/estimate the contact force. However, in such work, the authors neglect some dynamics as well as external disturbances, in addition to, the proposed algorithm is model-based and it does not have a robustness capability. In \cite{ruggiero2014impedance}, a scheme is proposed which allows a quadrotor to perform tracking tasks without a precise knowledge of its dynamics and under the effect of external disturbances and unmodeled aerodynamics. In addition, this scheme can estimate the external generalized forces. However, as the authors claim, this estimator can work perfectly with constant external disturbances. In addition, the estimated forces contain many different types of forces such as wind, payload, environment impacts, and unmodeled dynamics. Thus, it can not isolate the end-effector force only from the others. The authors in \cite{tomic2014unified} present a model-based method to estimate the external wrench of a flying robot. However, this method assumes that there are no modeling errors and no external disturbance. Moreover, it estimates the external force as one unite and it can not distinguish between external disturbance and the end-effector force which we need to calculate for teleoperation purposes. In addition, it uses a model based control which needs a full knowledge of the model.       

In this article, a new scheme is proposed to cope up with these limitations of the currently developed techniques to solve the issues of this complicated multibody robotic system. Firstly, a DOb inner loop is used to estimate both the system nonlinearities and all external forces to compensate for them, as a result, the system acts like a linear decoupled MIMO system. Secondly, a fast tracking RLS algorithm is utilized with the linearization capabilities of DOb to estimate the contact force, in addition to, it enables the user to sense the contact force at the end-effector that it is not available in the current developed schemes. Thirdly, a model-free robust impedance control of the quadrotor manipulation system is implemented. The DOb is designed in the quadrotor/joint space while the impedance control is designed in the task space such that the end-effector can track the desired task space trajectories besides applying a specified environment impedance. Thus, Fourthly, a Jacobian based algorithm is proposed to transform the control signal from the task space to the quadrotor/joint space coordinates. The rigorous stability analysis of the proposed scheme is presented. Finally, the system model is simulated in MATLAB taking in to considerations all the non-idealities and based on real parameters to emulate a real system. 

\section{System Modeling} \label{se:model}
\begin{figure}[!t]
	\centering
	\includegraphics[width=0.8\columnwidth, height = 5 cm]{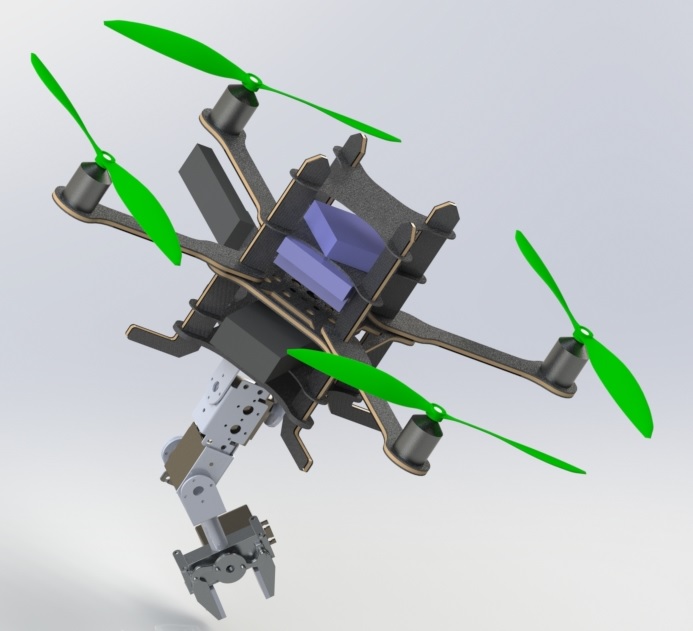}
	\caption{3D CAD model of the proposed quadrotor-based aerial manipulator}
	\label{3D-CAD-MODEL}
\end{figure}

\begin{figure}[!t]
	\centering
	\includegraphics[width=0.95\columnwidth]{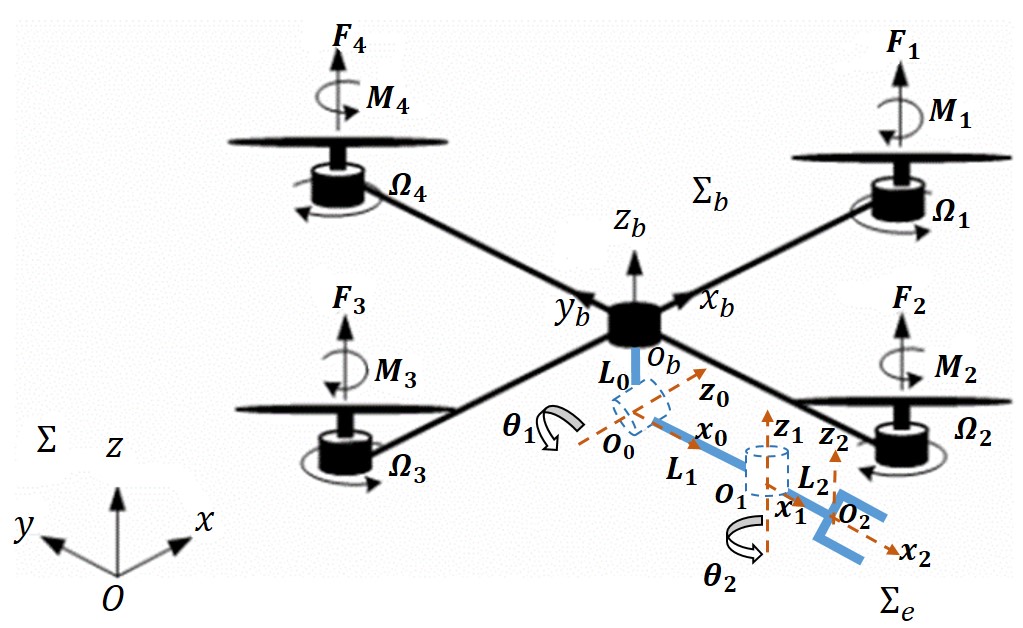}
	\caption{Quadrotor-based aerial manipulator with relevant frames}
	\label{frames}
\end{figure}

Fig. \ref{3D-CAD-MODEL} presents a 3D CAD model of the proposed quadrotor-based aerial manipulator. The system is composed of a manipulator mounted on the bottom center of a quadrotor.
System geometrical frames, which are assumed to satisfy the Denavit-Hartenberg (DH) convention, are illustrated in Fig. \ref{frames}. The manipulator has two revolute joints. The axis of the first revolute joint, $z_0$, is parallel to the quadrotor $x$-axis. The axis of the second joint, $z_1$, is normal to that of the first joint and hence it is parallel to the quadrotor $y$-axis at the extended configuration. Therefore, the pitching and rolling rotation of the end-effector is allowable independently from the horizontal motion of the quadrotor. Hence, with this proposed aerial manipulator, it is possible to manipulate objects with arbitrary location and orientation. Consequently, the end-effector can make motion in 6-DOF with minimum possible number of actuators/links that is critical factor in flight.

The quadrotor components are designed to achieve a payload capacity of $500$  g. Asctec pelican quadrotor \cite{asctec} is utilized as a quadrotor platform. The maximum thrust force for each rotor is $6$N. The arm is designed so that the total weight of the arm is $200$ g, it has a maximum reach in the range of $22$, and it can carry a payload of $200$  g. It has three DC motors, (HS-422 (Max torque = $0.4$  N.m) for gripper, HS-5485HB  (Max torque = $0.7$  N.m) for joint $1$, and HS-422 (Max torque = $0.4$  N.m) for joint $2$).

The angular velocity of each rotor $j$ is $\Omega_j$ and it generates thrust force $F_j$ and drag moment $M_j$ that are given by
\begin{equation}
F_j = K_{f_j} \Omega_j^2,
\label{thrust}
\end{equation}
\begin{equation}
M_j = K_{m_j} \Omega_j^2,
\label{dragmoment}
\end{equation}
where $K_{f_j}$ and $K_{m_j}$ are the thrust and drag coefficients.
\subsection{Kinematics}
Let $\Sigma_b$, $O_{b}$- $x_b$ $y_b$ $z_b$, represents the quadrotor body-fixed reference frame with origin at the quadrotor center of mass, see Fig. \ref{frames}. Its position with respect to the world-fixed inertial reference frame, $\Sigma$, $O$- $x$ $y$ $z$, is given by the $(3 \times 1)$ vector $p_b=[x, y, z]^T$, while its orientation is given by the rotation matrix $R_b$ which is given by
\begin{equation}
R_b= \begin{bmatrix}
C_{\psi} C_{\theta}  ~ ~ & ~ ~ S_{\phi} S_{\theta} C_{\psi}-S_{\psi} C_{\phi}  ~ ~ &  ~ ~ S_{\psi} S_{\phi}+C_{\psi} S_{\theta} C_{\phi} \\
S_{\psi} C_{\theta}  ~ ~ & ~ ~ C_{\psi} C_{\phi}+ S_{\psi} S_{\theta} S_{\phi}  ~ ~ &  ~ ~ S_{\psi} S_{\theta} C_{\phi}-C_{\psi} S_{\phi} \\
-S_{\theta}  ~ ~ & ~ ~ C_{\theta} S_{\phi}  ~ ~ &  ~ ~ C_{\theta} C_{\phi} \\
\end{bmatrix},
\label{eq:Rb}
\end{equation}
where $\Phi_b$=$[\psi,\theta,\phi]^T$ is the triple $ZYX$ yaw-pitch-roll angles. Note that $C(.)$ and $S(.)$ are short notations for $cos(.)$ and $sin(.)$.
Let us consider the frame  $\Sigma_e$, $O_{2}$- $x_2$ $y_2$ $z_2$, attached to the end-effector of the manipulator, see Fig. \ref{frames}. 
Thus, the position of $\Sigma_e$ with respect to $\Sigma$ is given by
\begin{equation}
p_e = p_b + R_b p^b_{eb},
\label{eq:pe}
\end{equation}
where the vector $p^b_{eb}$ describes the position of $\Sigma_e$ with respect to $\Sigma_b$ expressed in $\Sigma_b$. The orientation of $\Sigma_e$ can be defined by the rotation matrix
\begin{equation}
R_e = R_b R^b_e,
\label{eq:Re_cpct}
\end{equation}
where $R^b_e$ describes the orientation of $\Sigma_e$ w.r.t $\Sigma_b$. 
The linear velocity $\dot{p}_e$ of $\Sigma_e$ in the world-fixed frame is obtained by the differentiation of (\ref{eq:pe}) as
\begin{equation}
\dot{p}_e = \dot{p}_b - Skew(R_b p^b_{eb}) \omega_b + R_b \dot{p}^b_{eb},
\label{eq:pde}
\end{equation}
where $Skew(.)$ is the $(3 \times 3)$ skew-symmetric matrix operator \cite{spong2006robot}, while $\omega_b$ is the angular velocity of the quadrotor expressed in $\Sigma$.
The angular velocity $\omega_e$ of $\Sigma_e$ is expressed as
\begin{equation}
\omega_e = \omega_b + R_b \omega^b_{eb},
\label{eq:we}
\end{equation}
where $\omega^b_{eb}$ is the angular velocity of the end-effector relative to $\Sigma_b$ and is expressed in $\Sigma_b$.

Let $\Theta = [\theta_1, \theta_2]^T$ be the $(2 \times 1)$ vector of joint angles of the manipulator. The $(6 \times 1)$ vector of the generalized velocity of the end-effector with respect to $\Sigma_b$, $v^b_{eb} = [\dot{p}^{bT}_{eb},\omega^{bT}_{eb}]^T$, can be expressed in terms of the joint velocities $\dot{\Theta}$ via the manipulator Jacobian $J^b_{eb}$ \cite{Tsai}, such that
\begin{equation}
v^b_{eb} = J^b_{eb} \dot{\Theta}.
\label{eq:vbeb}
\end{equation}

From (\ref{eq:pde}) and (\ref{eq:we}), the generalized end-effector velocity, $v_e = [\dot{p}^T_e, \omega^T_e]^T$, can be expressed as

\begin{equation}
v_e = J_b v_b + J_{eb} \dot{\Theta},
\label{eq:ve}
\end{equation}
where $v_b = [\dot{p}^T_b,\omega^T_b]^T$,
$J_b= \begin{bmatrix}
I_3 & -Skew(R_b p^b_{eb})\\
O_3 & I_3
\end{bmatrix},$
$J_{eb}= \begin{bmatrix}
R_b & O_3\\
O_3 & R_b
\end{bmatrix} J^b_{eb}$,\\ 
where $I_m$ and $O_m$ denote $(m \times m)$ identity and $(m \times m)$ null matrices, respectively.
If the attitude of the vehicle is expressed in terms of yaw-pitch-roll angles, then (\ref{eq:ve}) will be
\begin{equation}
v_e = J_b Q_b \chi_b + J_{eb} \dot{\Theta},
\label{eq:veph}
\end{equation}
with 
$\chi_b= \begin{bmatrix}
p_b \\
\Phi_b 
\end{bmatrix},$ 
$Q_b= \begin{bmatrix}
I_3 & O_3\\
O_3 & T_b
\end{bmatrix},$ 
where $T_b$ describes the transformation matrix between the angular velocity $\omega_b$ and the time derivative of Euler angles $\dot{\Phi}_b$, and it is given as
\begin{equation}
T_b(\Phi_b)= \begin{bmatrix}
0 &  -S(\psi) & C(\psi) C(\theta) \\
0 & C(\psi) & S(\psi) C(\theta) \\
1 & 0 & -S(\theta) \\
\end{bmatrix}.
\label{eq:Tb}
\end{equation}

Since the vehicle is an under-actuated system, i.e., only $4$ independent control inputs are available for the 6-DOF system, the position and the yaw angle are usually the controlled variables. The pitch and roll angles are used as intermediate control inputs to control the horizontal position. Hence, it is worth rewriting the vector $\chi_b$ as follows
$
\chi_b= \begin{bmatrix}
\eta_b \\
\sigma_b 
\end{bmatrix}, 
$
$
\eta_b= \begin{bmatrix}
p_b \\
\psi 
\end{bmatrix}, 
$  
$
\sigma_b= \begin{bmatrix}
\theta \\
\phi 
\end{bmatrix}. 
$ 

Thus, the differential kinematics (\ref{eq:veph}) will be
\begin{equation}
\begin{aligned}
v_e &= J_{\eta} \dot{\eta}_b + J_{\sigma} \dot{\sigma}_b + J_{eb} \dot{\Theta}\\
&=J_{\zeta} \dot{\zeta} + J_{\sigma} \dot{\sigma}_b,
\end{aligned} 
\label{eq:vediv}
\end{equation}
where $\zeta = [\eta_b^T,\Theta^T]^T$ is the vector of the controlled variables, $J_{\eta}$ is composed by the first 4 columns of $J_b Q_b$, $J_{\sigma}$ is composed by the last 2 columns of $J_b Q_b$, and $J_{\zeta} = [J_{\eta}, J_{eb}]$.

If the end-effector orientation is expressed via a triple of Euler angles, $ZYX$, $\Phi_e$, the differential kinematics (\ref{eq:vediv}) can be rewritten in terms of the vector $\dot{\chi}_e = [\dot{p}^T_e, \dot{\Phi}_e^T]^T$ as follows
\begin{equation}
\begin{aligned}
\dot{\chi}_e &= Q_e^{-1}(\Phi_e) v_e\\
&=Q_e^{-1}(\Phi_e) [J_{\zeta} \dot{\zeta} + J_{\sigma} \dot{\sigma}_b],
\end{aligned} 
\label{eq:xedot}
\end{equation}
where $Q_e$ is the same as $Q_b$ but it is a function of $\Phi_e$ instead of $\Phi_b$.

\subsection{Dynamics}
The equations of motion of the proposed robot have been derived in details in \cite{new_quad_manp}. The dynamical model of the quadrotor-manipulator system can be reformulated in a matrix form as
\begin{equation}
M(q) \ddot{q} + C(q,\dot{q}) \dot{q} + G(q) + \tau_w + \tau_l=\tau, \qquad \tau = B u,
\label{eq:dyn_gen}
\end{equation}
where $q=[x, y, z, \psi, \theta, \phi, \theta_1, \theta_2]^T$  $ \in R^{8}$ represents the vector of the generalized coordinates, $M$ $ \in R^{8 \times 8}$ denotes the symmetric and positive definite inertia matrix of the system, $C$ $ \in R^{8 \times 8}$ represents the Coriolis and centrifugal terms, $G$ $ \in R^{8}$ represents the gravity term, $\tau_w $ $ \in R^{8}$ is vector of the external disturbances, $\tau_l $ $ \in R^{8}$ is vector of the contact force effect, $\tau$ $ \in R^{8}$ is the generalized input torques/forces, $u = [F_1, F_2, F_3, F_4, \tau_{m_1}, \tau_{m_2}]^T$ is vector of the actuator inputs, and $B= H N$ is the input matrix which is used to produced the body forces and moments from the actuator inputs. The control matrix,
$N$, is given as
\begin{equation}
N= \begin{bmatrix}
0 & 0 & 0 & 0 & 0 & 0 \\
0 & 0 & 0 & 0 & 0 & 0 \\
1 & 1 & 1 & 1 & 0 & 0 \\
\gamma_1 & -\gamma_2 & \gamma_3 & -\gamma_4 & 0 & 0 \\
-d & 0 & d & 0 & 0 & 0 \\
0 & -d & 0 & d & 0 & 0 \\
0 & 0 & 0 & 0 & 1 & 0 \\
0 & 0 & 0 & 0 & 0 & 1 
\end{bmatrix},
\label{eq:N}
\end{equation}

where $\gamma_j=K_{m_j}/K_{f_j}$, and $H$ $ \in R^{8 \times 8}$ is matrix that transforms body input forces to be expressed in $\Sigma$ and is given by
\begin{equation}
H= \begin{bmatrix}
R_b & O_3 & O_2 \\
O_3 & T_b^T R_b & O_2 \\
O_{2x3} & O_{2x3} & I_2 
\end{bmatrix}.
\label{eq:H}
\end{equation}
The environment dynamics, contact force, $\tau_l$, can be modeled as following:
\begin{equation}
\begin{split}
\tau_l = J^T F_e,\\
F_e= S_c \chi_e + D_c \dot{\chi}_e,
\end{split}
\label{eq:Tl}
\end{equation}
where $S_c = diag\{S_{c_1}, S_{c_2}, S_{c_3}, S_{c_4}, S_{c_5}, S_{c_6}\}$ and $D_c = diag\{D_{c_1}, D_{c_2}, D_{c_3}, D_{c_4}, D_{c_5}, D_{c_6}\}$ represent the environment stiffness and the environment damping, receptively.

The wind dynamics, $\tau_w$, can be modeled as following \cite{windmodel, hsu2011verifying, andrews2012modeling}:

The average wind velocity is determined by 
\begin{equation}
V_{wz} = V_{w_{z_0}} \frac{z}{z_0},
\label{eq:Vwz}
\end{equation}
where $V_{wz}$  is the wind velocity at altitude $z$, $V_{w_{z_0}}$ is the specified 
(measured) wind velocity at altitude $z_0$.
To  simulate  wind  disturbances,  it is worth calculating the wind force, $F_{w}$, which influences the platform than the wind  velocity.  This  force  can  be  determined  by
\begin{equation}
F_{w} = 0.61 * A_e V_{wz}^2, 
\label{eq:Fw}
\end{equation}
where $0.61$ is used to convert wind velocity to pressure, and $A_e$  is the influence effective area which depends on the quadrotor structure and its orientation.

This force can be projected on the axes of frame $\Sigma$ as
\begin{equation}
\begin{split}
F_{wx} = f_{wx_1} z^2 sin(\theta) + f_{wx_2} z^2 cos(\theta), \\
F_{wy} = f_{wy_1} z^2 sin(\phi) + f_{wy_2} z^2 cos(\phi),
\end{split}
\label{eq:Fwxy}
\end{equation} 
where $f_{wx_1} = 0.61 * A_{e_1} (\frac{V_{w_{z_0}}}{z_0})^2 cos(\psi_w)$, $f_{wx_2} = 0.61 * A_{e_2} (\frac{V_{w_{z_0}}}{z_0})^2 cos(\psi_w)$, $f_{wy_1} = 0.61 * A_{e_1} (\frac{V_{w_{z_0}}}{z_0})^2 sin(\psi_w)$, $f_{wy_2} = 0.61 * A_{e_2} (\frac{V_{w_{z_0}}}{z_0})^2 sin(\psi_w)$, $\psi_w$ represents the angle of wind direction, and both $A_{e_1}$ and $A_{e_2}$ depend on the quadrotor design parameters.

\section{Controller Design} \label{se:control}

\subsection{Control Objectives} \label{sse:cntrl_obj}
\begin{figure}[!t]
	\centering
	\includegraphics[width=0.95\columnwidth, height = 5 cm]{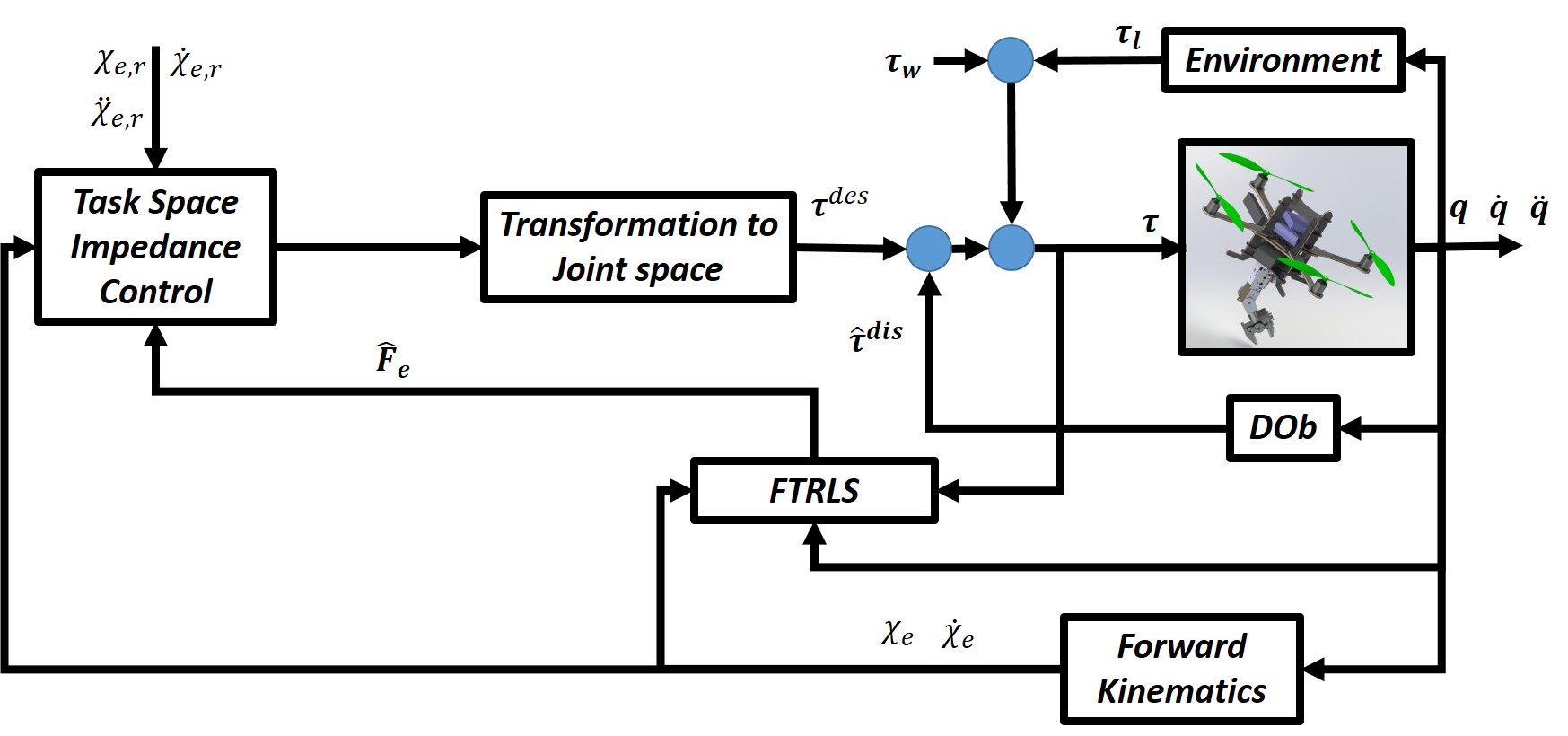}
	\caption{Functional block diagram of the proposed control scheme}
	\label{fig:mpc_dob_fncblg}
\end{figure}

Our goal is to design of estimation and control system to achieve the following objectives:
\begin{enumerate}
	\item Robust Stability: The robotic system in Fig. \ref{fig:mpc_dob_fncblg}  is robust and stable against the external disturbances, parameters uncertainties, and noises.
	\item Force Estimation: The end-effector contact force has to be estimated with fast response and the estimation error tends to zero as the time tends to $\infty$.
	\item $6$-DOF Impedance Control: In the presence of the applied force/desired impedance at the end-effector, the end-effector tracking error tends to zero as time tends to $\infty$.
\end{enumerate}

To this end, we propose a control scheme as shown in Fig. \ref{fig:mpc_dob_fncblg}. In this control strategy, the system nonlinearities, external disturbances (wind), $\tau_{w}$, and contact force, $\tau_{l}$, are treated as disturbances, $\tau^{dis}$, that will be estimated, $\hat{\tau}^{dis}$, and compensated by the DOb in the inner loop. The system can be now tackled as linear SISO plants. The output of DOb with system measurements of both joint and task spaces variables are used as the inputs to the FTRLS to obtain the end-effector contact force $\hat{F}_e$. The task space impedance control is used in the external loop of DOb and its output is transformed to the joint space through a transformation algorithm.
\subsection{Disturbance Observer Loop} \label{sse:dob}
\begin{figure}[!t]
	\centering
	\includegraphics[width=0.95\columnwidth, height = 5 cm]{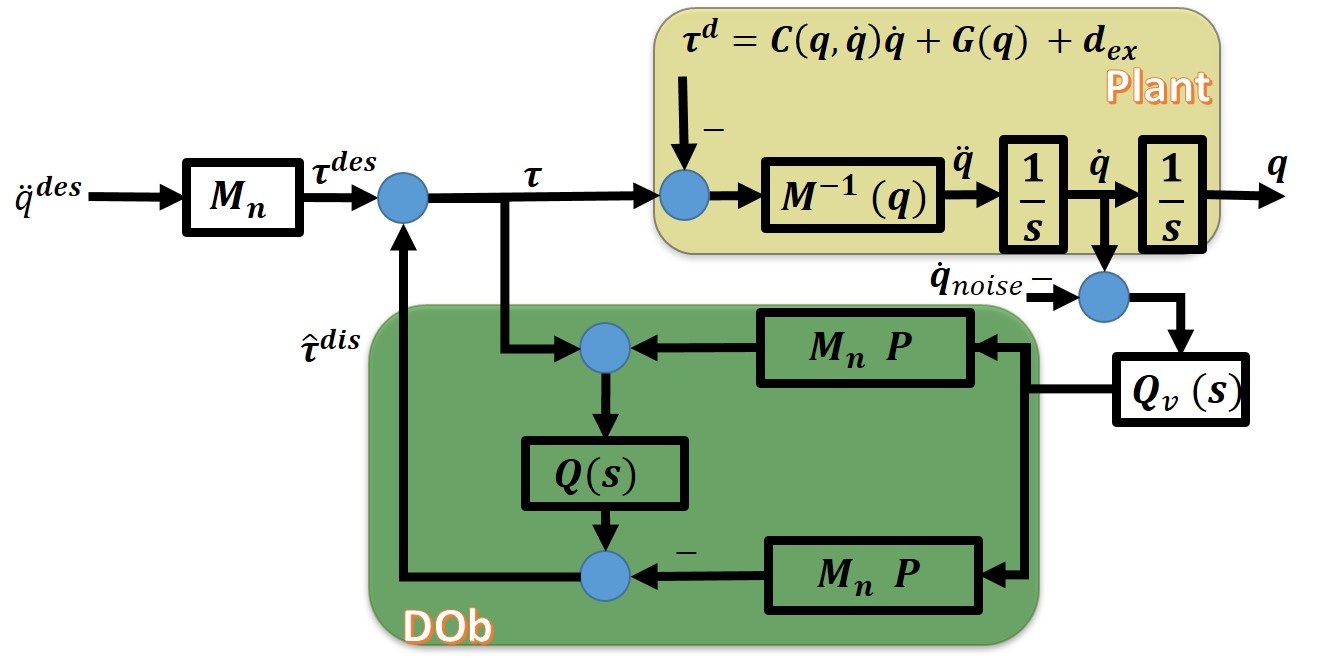}
	\caption{Block diagram of DOb internal loop}
	\label{fig:interlp_DOb}
\end{figure}

A block diagram of the DOb inner loop is shown in Fig. \ref{fig:interlp_DOb}. In this figure, $M_n$ $\in R^{8 \times 8}$ is the system nominal inertia matrix, $\tau$ and $\tau^{des}$ are the robot and desired inputs, respectively, $P=diag([g_1,...,g_i,...,g_8])$ with $g_i$ is the bandwidth of the $i^{th}$ variable of $q$, $Q(s)=diag([\frac{g_1}{s+g_1} ,...,\frac{g_i}{s+g_i},...,\frac{g_8}{s+g_8}])$ $\in R^{8 \times 8}$ is the matrix of the low pass filter of DOb. The DOb requires velocity measurement. Practically, the velocity have to be fed through a low pass filter, $Q_v(s)=diag([\frac{g_{v_1}}{s+g_{v_1}},...,\frac{g_{v_i}}{s+g_{v_i}},...,\frac{g_{v_8}}{s+g_{v_8}}])$ $\in R^{8 \times 8}$, and with cut-off frequency of $P_v=diag([g_{v_1},...,g_{v_i},...,g_{v_8}])$. $\tau^{dis}$ represents the system disturbances, and $\hat{\tau}^{dis}$ is the estimated disturbances. 

If  we  apply  the  concept  of  disturbance  observer to the proposed system, the independent coordinate  control is possible  without  considering  coupling effect of other coordinates. The  coupling  terms  such  as centripetal  and  Coriolis and gravity terms are considered as disturbance and compensated by feed forward the estimated disturbance torque.  

The disturbance $\tau^{dis}$ can be assumed as
\begin{equation}
\begin{split}
\tau^{dis} = (M(q) - M_n) \ddot{q} + \tau^d,\\ 
\tau^d=C(q,\dot{q}) \dot{q} + G(q) + d_{ex}.
\label{eq:sys_dis}
\end{split}
\end{equation}  
Substituting from (\ref{eq:sys_dis}), then (\ref{eq:dyn_gen}) can be rewritten as
\begin{equation}
M_n \ddot{q} + \tau^{dis} = \tau. 
\label{eq:sys_eq_new}
\end{equation}
The control input, $\tau$, see Fig. \ref{fig:interlp_DOb}, is given as
\begin{equation}
\begin{split}
\tau = \frac{1}{(1-Q(s))}[M_n \ddot{q}^{des} - Q(s) M_n \ddot{q}], \\ = M_n \ddot{q}^{des} + M_n P e_v, \quad e_v = \dot{q}^{des} - \dot{q}.
\label{eq:tau}
\end{split}
\end{equation}  
Applying this control law results in
\begin{equation}
\begin{split}
M(q) \dot{e}_v + C(q,\dot{q}) e_v +  K_v e_v = \delta, 
\\ K_v = P M_n,
\label{eq:dyn_gen_dobeD}
\end{split} 
\end{equation} 
where
\begin{equation}
\begin{split}
\delta = \Delta M(q) \ddot{q}^{des} + C(q,\dot{q}) \dot{q}^{des} + G(q) + d_{ex}, \\ \Delta M(q) = M(q) - M_n.
\label{eq:delta}
\end{split}
\end{equation}

Stability of this inner loop can be proved as following:

To simplify the analysis, let us ignore the effect of the velocity filter which will be considered later. 

Let us use a Lyapunov function as
\begin{equation}
V=\frac{1}{2} e_v^T M(q) e_v.
\label{eq:lyp_fun}
\end{equation} 
The time derivative of this function is
\begin{equation}
\dot{V}= e_v^T M(q) \dot{e}_v + \frac{1}{2} e_v^T \dot{M}(q) e_v.
\label{eq:lyp_der1}
\end{equation} 
Substituting from (\ref{eq:dyn_gen_dobeD}), then (\ref{eq:lyp_der1}) becomes
\begin{equation}
\dot{V}= e_v^T \delta - e_v^T K_v e_v + \frac{1}{2} e_v^T (\dot{M}(q) - 2 C(q,\dot{q}))e_v. 
\label{eq:lyp_der2}
\end{equation} 

To complete this proof, the properties of the dynamic equation of motion (\ref{eq:dyn_gen}) will be utilized. Theses properties are  \cite{from2014vehicle, spong2006robot}:

\begin{property}
	\begin{equation}
	\lambda_{min} \norm{\nu}^2  \leq  \nu^T M(q) \nu \leq \lambda_{max} \norm{\nu}^2, 
	\label{eq:M_pro}
	\end{equation}
\end{property}

\begin{property}
	\begin{equation}
	\nu^T(\dot{M}(q) - 2 C(q,\dot{q}))\nu=0,
	\label{eq:d/dt_pro}
	\end{equation} 
\end{property}
where $\nu \in R^8$ represents a $8$-dimensional vector, and $\lambda_{min}$ and $\lambda_{max}$ are positive real constants. 

Substituting from (\ref{eq:d/dt_pro}), then (\ref{eq:lyp_der2}) will be
\begin{equation}
\dot{V}= e_v^T \delta - e_v^T K_v e_v. 
\label{eq:lyp_der3}
\end{equation}
From property (\ref{eq:M_pro}), one can get
\begin{equation}
\dot{V} \leq -\gamma V + \sqrt{\frac{2 V}{\lambda_{min}}} |\delta|, \quad 
\gamma = \frac{2K_v}{\lambda_{max}}. 
\label{eq:lyp_der4}
\end{equation}
From the analysis presented in \cite{sadegh1990stability}, (\ref{eq:lyp_der4}) can be reformulated as
\begin{equation}
\norm{e_v}_p \leq \frac{1}{\gamma} + \sqrt{\frac{2}{\lambda_{min}}} (\frac{2}{p\gamma})^{\frac{1}{p}} \sqrt{V(0,e_v(0))} \norm{\delta}_p.
\label{eq:ev_final}
\end{equation}
Thus, the error dynamics is $L_p$ input/output stable with respect to the pair ($\delta$,$e_v$) for all $p \in [1,\infty]$ with the assumption that the system states, $q$ and $\dot{q}$, are bounded.

If one considers the effect of using the velocity filter, then the characteristic equation of the inner loop is
\begin{equation}
P_{c_i} = s^2 + g_{v_i} s + \alpha_i g_i g_{v_i},
\label{eq:ch_eq}
\end{equation}
where $\alpha_i = \frac{M_{n_{ii}}}{M_{ii}}$.

To improve the robustness, the damping coefficient of this equation, which is $ 0.5 \sqrt{\frac{g_i g_{v_i}}{\alpha g_i}}$, should larger than or equal $0.707$ . Therefore, the following inequality
\begin{equation}
\alpha g_i \leq \frac{g_{v_i}}{2},
\label{eq:gi_gv}
\end{equation}
should be hold.
Recasting (\ref{eq:gi_gv}) with respect to $K_v$ gives to
\begin{equation}
\frac{K_{v_i} }{M_{ii}}\leq \frac{g_{v_i}}{2}. 
\label{eq:gi_gv_Kv}
\end{equation}

Summarizing, (\ref{eq:lyp_der4}) shows that the stability and robustness of the control system is enhanced by increasing $K_v$, i.e., by increasing $M_n$ and $P$, but without violating the robustness constraint given in (\ref{eq:gi_gv_Kv}).

If the DOb performs well, that is $\hat{\tau}^{dis}$ = $\tau^{dis}$, the dynamics from the DOb loop input $\tau^{des}$ to the output of the system is given as 
\begin{equation}
M_n \ddot{q}=\tau^{des}.
\label{eq:sys_eq_reduced}
\end{equation}
Since $M_n$ is assumed to be a diagonal matrix, the system can be considered as a decoupled linear multi SISO systems as
\begin{equation}
M_{n_{ii}} \ddot{q}_i=\tau_i^{des},
\label{eq:sys_eq_decoupled}
\end{equation}
or in the acceleration space as:
\begin{equation}
\ddot{q}_i=\ddot{q}_i^{des}.
\label{eq:sys_eq_decoupled_acc}
\end{equation}

The next step is to design an Impedance tracking based controller in the outer loop for the system of (\ref{eq:sys_eq_decoupled_acc}).
\subsection{Fast Tracking Recursive Least Squares} \label{sse:sprls}
In this part, we develop a technique which utilizes a Fast Tracking Recursive Least Squares (FTRLS) to estimate the contact force with the aid of the DOb linearization capabilities. The FTRLS algorithm is one of the fast online least squares-based identification methods used for the identification of environments with varying dynamic parameters \cite{hu2014least, wang2015recursive}. To apply FTRLS, the dynamic equations (\ref{eq:dyn_gen}- \ref{eq:Fwxy}) have to be parametrized (i.e., to be product of measurement data regressor and dynamic parameters) as follows:

The system dynamic part, $\tau_{int}=  M(q) \ddot{q} + C(q,\dot{q}) \dot{q} + G(q)$, can be rewritten as the product of data regressor, $Y_i(q,\dot{q},\ddot{q})$, and platform parameters, $h_i$. The environment dynamics, $\tau_l$, can be reformulated as $Y_l(q,\dot{q},\ddot{q},\chi_e,\dot{\chi}_e)*h_l$, where, $Y_l=J^T Y_e$, $Y_e$ is a function of the end effector states, ($\chi_e$,$\dot{\chi}_e$), and $h_l$ is the environment parameters $S_c$ and $D_c$. Finally, the wind effect is formulated as $Y_w(z,\theta,\phi) * h_w$, where $h_w$ is the wind parameters. Thus, the total dynamics can be reformulated as
\begin{equation}
\begin{split}
\tau = Y * h, \\
Y=[Y_i, Y_l, Y_w], \\
h=[h_i, h_l, h_w]^T,
\label{eq:reg_tot}
\end{split}
\end{equation}   
where $Y$ $\in R^{8 \times 40}$ and $h$ $\in R^{40}$ are the data regressor and parameters vector of (\ref{eq:dyn_gen}), respectively.

The parameter estimation error is
\begin{equation}
\tilde{h}(t) = h - \hat{h}(t),
\label{eq:SPRLS_eh}
\end{equation}
while the estimation error is
\begin{equation}
\tilde{\tau}(t) = \tau(t) - Y(t) \hat{h}(t) = Y(t) \tilde{h}(t).
\label{eq:SPRLS_ey}
\end{equation}
By minimizing a cost function with respect to the parameter estimation error, one can find the time derivative of the estimated parameters vector, $\hat{h}$, as following
\begin{equation}
\frac{d}{dt}\hat{h}(t)=  R(t) Y^{T}(t) \tilde{\tau}(t), 
\label{eq:SPRLS_hd}
\end{equation}
where $R(t)$ is the parameters' covariance matrix, and it can be calculated from
\begin{equation}
\frac{d}{dt}R^{-1}(t) = -\eta_h(t) R^{-1}(t) + Y^{T}(t) Y(t),
\label{eq:SPRLS_R}
\end{equation}
where $\eta_h$ is the forgetting factor, and it is given as 
\begin{equation}
\eta_h(t) = \eta_h^{min}  + (1-\eta_h^{min}) 2^{(-NINT(\gamma_g \norm{\tilde{\tau}(t)}^2))},
\label{eq:SPRLS_Rg}
\end{equation}
where $\eta_h^{min}$ is a constant representing the minimum forgetting factor, $NINT(.)$ is the round-off operator, and $\gamma_g$ is a design constant. This adaptive formulation of the forgetting factor enables the RLS to track the non-stationary parameters to be estimated.  

The convergence/stability ($\tilde{h}(t) \longrightarrow 0$) proof of this algorithm can be implemented as following:

Let us assume the Lyapunov function as
\begin{equation}
V(t) = \tilde{h}^T(t) R^{-1}(t) \tilde{h}(t).
\label{eq:SPRLS_lyap}
\end{equation}
If $R^{-1}(t)$ is chosen to be positive definite, then $V(t)$ will be positive definite. To prove the positive definiteness of $R^{-1}(t)$, let us use the solution of the differential equation (\ref{eq:SPRLS_R}) which is
\begin{multline}
R^{-1}(t) = \Phi_h(t,t_0) R^{-1}(t_0) \Phi_h^T(t,t_0) + \\ \int_{t_0}^{t} \Phi_h(t,\varrho) Y^T(\varrho) Y(\varrho) \Phi_h^T(t,\varrho) d\varrho,  
\label{eq:SPRLS_R_sol}
\end{multline}
where $\Phi_h^T(t,t_0)$ is the state transition matrix of a system described by $\dot{\upsilon}(t) = - \frac{1}{2}\eta_h \upsilon(t)$. 
Thus, by choosing $R^{-1}(t_0) > 0$, then the first term in (\ref{eq:SPRLS_R_sol}) will be positive definite.
The second term is also positive definite. As a result, the proposed covariance matrix update formula is positive definite, and thus, the chosen Lyapunov function (\ref{eq:SPRLS_lyap}) is positive definite.

The time derivative of Lyapunov function is
\begin{equation}
\dot{V}(t) = 2 \tilde{h}^T R^{-1} \dot{\tilde{h}} +  \tilde{h}^T \dot{R^{-1}} \tilde{h}.
\label{eq:SPRLS_lyapd}
\end{equation}   
However, by differentiating both sides of (\ref{eq:SPRLS_eh}) with respect to time, one can find that $\dot{\tilde{h}} = - \dot{\hat{h}}$, by substituting from the proposed formula of $\dot{\hat{h}}$ (\ref{eq:SPRLS_hd}) and (\ref{eq:SPRLS_ey}), then
\begin{equation}
\dot{\tilde{h}} = - R Y^T Y \tilde{h}.  
\label{eq:SPRLS_lyapd2}
\end{equation}  
Substituting from (\ref{eq:SPRLS_lyapd2}) in (\ref{eq:SPRLS_lyapd}), then $\dot{V}(t)$ will be
\begin{equation}
\dot{V}(t) = -\tilde{h}^T [2 Y^T Y - \dot{R^{-1}}] \tilde{h}.
\label{eq:SPRLS_lyapd3}
\end{equation}   
Substituting from the proposed formula (\ref{eq:SPRLS_R}) for $\dot{R^{-1}}$ into (\ref{eq:SPRLS_lyapd3}), then
\begin{equation}
\dot{V}(t) = -\tilde{h}^T [Y^T Y +\eta_h(t) R^{-1}(t)] \tilde{h}.
\label{eq:SPRLS_lyapd4}
\end{equation} 
Thus, the time derivative of $V(t)$ is negative definite which ensures the asymptotic stability of the estimation error ($\tilde{h}(t) \longrightarrow 0$ as $t \longrightarrow \infty$) 

Finally, for both teleoperation impedance control purposes, the user can calculate the estimated environment impedance, contact force, from  
\begin{equation}
\begin{split}
\hat{\tau}_l = Y_l \hat{h}_l,\\
\hat{F}_e = Y_e \hat{h}_l.
\end{split}
\label{eq:SPRLS_thl}
\end{equation}
Therefore, unlike the current developed schemes, with this technique, one can isolate and estimate the end-effector contact force apart from the whole estimated forces in the systems.

\subsection{Impedance Control} \label{sse:impd_ctrl}
The objective of the impedance control is to regulate the end-effector interaction force,  
which may vary due to the uncertainty in the location of the interaction point and/or the structural properties of the environment, besides achieving task space trajectory tracking.
The linear impedance control is designed in the task space. This is based on the linearization effect of the designed DOb in the joint space.
The desired acceleration in the task space, $\ddot{\chi}^{des}_e$, can be calculated from
\begin{equation}
\ddot{\chi}^{des}_e = \ddot{\chi}_{e,r} + S_{c,d} (\chi_{e,r} -  \chi_{e}) + D_{c,d} (\dot{\chi}_{e,r} -  \dot{\chi}_{e}) - \hat{F}_e,
\label{eq:xe_des}
\end{equation}
where $S_{c,d}$ and $D_{c,d}$ are the desired values of $S_{c}$ and $D_{c}$ respectively, which determine the desired impedance that the end-effector will apply to the environment.
\textcolor{black}{
	Let us define the quadrotor/joint space tracking error as
	\begin{gather}
		e = q_r - q, \quad \dot{e} = \dot{q}_r - \dot{q}, \quad
		\ddot{e} = \ddot{q}_r - \ddot{q}, 
	\end{gather}
	while the task space tracking error can be defined as
	\begin{gather}
		e_e = \chi_{e,r} - \chi_{e}, \quad \dot{e}_e = \dot{\chi}_{e,r} - \dot{\chi}_{e},
		\quad \ddot{e}_e = \ddot{\chi}_{e,r} - \ddot{\chi}_{e}, 
	\end{gather} 
	where $\chi_{e,r}$, $\dot{\chi}_{e,r}$, and $\ddot{\chi}_{e,r}$ are the reference trajectories for the position, velocity, acceleration in the task space, respectively which are chosen to be bounded and continuous. $q_r$, $\dot{q}_{r}$, and $\ddot{q}_{r}$ are the reference trajectories for the position, velocity, acceleration in the quadrotor/joint space, respectively. Transformation from the task space to quadrotor/joint space will be implemented via the inverse of system Jacobian. 
}\textcolor{black}{
	The relation between the inner loop and the outer loop errors can be obtained as follows.
} 
\textcolor{black}{
	The DOb loop error can be expressed in the task space, $e_{v,e}$, via the Jacobian by
	\begin{equation}
	e_{v,e} = J e_v,
	\label{eq:ev_task}
	\end{equation}
	where $e_{v,e} = \dot{\chi}_e^{des} - \dot{\chi}_e$. From the previous analysis, it is proved that $e_v$ is bounded as in (\ref{eq:ev_final}). If we define $\dot{e}_{v,e} = \ddot{\chi}_e^{des} - \ddot{\chi}_e$, then by substituting from (\ref{eq:xe_des}), one can get
	\begin{equation}
	\dot{e}_{v,e} = \ddot{e}_e + S_{c,d} e_e + D_{c,d} \dot{e}_e - \hat{F}_e.
	\label{eq:eve}
	\end{equation} 
	Equation (\ref{eq:eve}) can be reformulated in a state space form as
	\begin{equation}
	\dot{X}_e = A_e X_e + B_e U_e,
	\label{eq:eve_ss}
	\end{equation}
	where $X_e = \begin{bmatrix} e_e\\ \dot{e}_e\\ \end{bmatrix}$, $A_e = \begin{bmatrix} O_6 & I_6\\ - S_{c,d} & -D_{c,d} \end{bmatrix}$, $B_e = \begin{bmatrix} O_6\\ I_6\\ \end{bmatrix}$, and $U_e = \dot{e}_{v,e} + \hat{F}_e$.   
} \textcolor{black}{
	By inspecting the matrix $A_e$ and based on the boundedness of both $\dot{e}_{v,e}$ and $\hat{F}_e$, one can find that the state, $X_e = [ e_e^T,  \dot{e}_e^T]^T$, is bounded and exponentially tends to zero as time tends to infinity as soon as the matrices, $S_{c,d}$ and $D_{c,d}$, are positive definite. As a result, since the Jacobian inverse exists (no singularities), the system errors, $e$ and $\dot{e}$, are also bounded and exponentially tends to zero as time tends to infinity. 
}

\begin{figure*}[!t]
	\centering
	\includegraphics[width=0.9\textwidth]{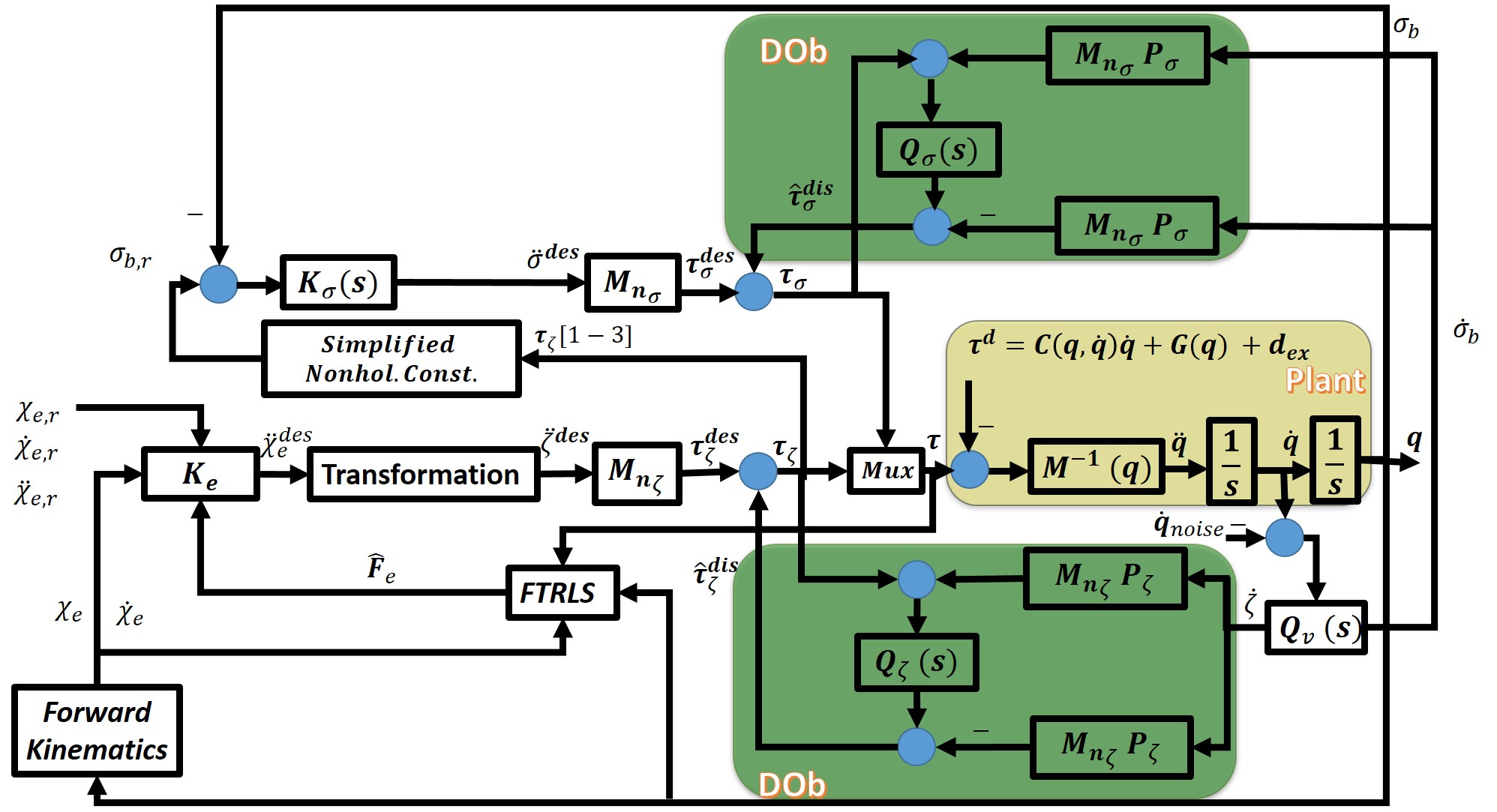}
	\caption{Details of the proposed control system}
	\label{fig:detail_impd_dob}
\end{figure*}

A complete and detailed  block diagram of the proposed control scheme is illustrated in Fig. \ref{fig:detail_impd_dob}. Quadrotor position and yaw rotation are the controlled variables, while pitch and roll angles are used as intermediate control inputs to achieve the desired $x$ and $y$. Therefore, the proposed scheme has two DOb-based controllers include one for $\zeta=[x, y, z, \psi, \theta_1, \theta_2]^T$ (with $M_{n_\zeta}$, $P_{\zeta}$, $Q_{\zeta}$) and the other for $\sigma_b = [\theta, \phi]^T$ (with $M_{n_\sigma}$, $P_{\sigma}$, $Q_{\sigma}$). The desired 6-DOF trajectories for the end-effector's ($\chi_{e,r}$), their actual values calculated by the forward kinematics, and the estimated end-effector force, are applied to the impedance control algorithm, $K_e$ that is given in (\ref{eq:xe_des}). Then, a transformation from task space to joint space is done by using (\ref{eq:qdd_des}) to get $\ddot{\zeta}^{des}$. The desired acceleration in the joint space, $\ddot{\zeta}^{des}$, can be calculated by differentiating (\ref{eq:xedot}) with respect to time as
\begin{equation}
\ddot{\zeta}^{des} = J_{\zeta}^{-1} (Q_e \ddot{\chi}^{des}_e + \dot{Q}_e \dot{\chi}_{e,r} - \dot{J}_{\zeta} \dot{\zeta} - J_{\sigma} \ddot{\sigma}_b - \dot{J}_{\sigma} \dot{\sigma}_b), 
\label{eq:qdd_des}
\end{equation}

The desired acceleration in quadrotor/joint space, $\ddot{\zeta}^{des}$, is then applied to the DOb of the independent coordinates, $\zeta$, to produce $\tau_{\zeta}$. The desired values for the intermediate DOb controller, $\sigma_{b,r}$, are obtained from the output of position controller, $\tau_{\zeta}$, through the following simplified nonholonomic constraints relation
\begin{equation}
\sigma_{b,r} = \frac{1}{\tau_{\zeta}(3)}
\begin{bmatrix}
C(\psi) & S(\psi) \\
S(\psi) & -C(\psi) 
\end{bmatrix}
\begin{bmatrix}
\tau_{\zeta}(1) \\
\tau_{\zeta}(2) 
\end{bmatrix}. 
\label{eq:sigmad}
\end{equation} 
The external controller of second DOb controller, $\tau_{\sigma}$, is used as a PD controller with velocity feedback, $K_{\sigma}$, as following:
\begin{equation}
\ddot{\sigma}^{des} = K_{p_{\sigma}} (\sigma_{b,r} - \sigma_{b}) - K_{d_{\sigma}} \dot{\sigma}_{b}.
\label{eq:K_sig}
\end{equation}
After that, $\ddot{\sigma}^{des}$ is applied to the second DOb to generate $\tau_{\sigma}$.

It is konwn that the response of the $\sigma$ controller must be much faster than that of the position controller. This can be achieved by the tuning parameters of both DOb and PD of $\sigma$-controller.  

The output of two controllers are converted to the required forces/torques applied to quadrotor/manipulator by
\begin{equation}
u= B_6^{-1} \begin{bmatrix}
\tau_{\zeta}(3,4)\\
\tau_{\sigma}\\
\tau_{\zeta}(5,6) 
\end{bmatrix},  
\label{eq:ufinal}
\end{equation}
where $B_6$ $\in R^{6 \times 6}$ is part of $B$ matrix and it is given by $B_6 = B(3:8,1:6)$.

\section{Simulation Results} \label{se:sim}

In this section, the presented aerial manipulation robot model with the proposed control technique is implemented in MATLAB/SIMULINK.
\subsection{Simulation Environment}
For a more realistic simulation studies, the following setup have been made:
\begin{itemize}
	\item - Linear and angular position and orientation of the quadrotor are available at rate of $1$ KHz. In \cite{achtelik2011onboard}, a scheme is proposed to measure and estimate the vehicle (Asctec Pelican Quadrotor) states based on IMU and Onboard camera in both indoors and outdoors.
	\item - The joints angles are measured at rate of $1$ KHz and angular velocities are estimated by a low pass filter.
	\item - The measured signal are affected by a normally distributed measurement noise with mean of $10^{-3}$ and standard deviation of $5 \times 10^{-3}$.
	\item - $1$ KHz Control loop.
	\item  - To test the robustness against model uncertainties, a step disturbance is applied at $15$ s to both the inertia matrix, $M(q)$, and the control matrix, $N$, (Actuators' losses) with $10 \% $ error.
\end{itemize}

\begin{table}[!t]
	\caption{System Parameters}
	\label{sys_par}
	\begin{center}
		\setlength{\tabcolsep}{0.001pt}
		\begin{tabular}{|c|c|c|c|c|c|}
			\hline
			Par.	&Value	&Unit&	Par. &	Value&	Unit \\
			\hline
			$m$&	$1$ &	$kg$ &	$L_2$&	$85\times10^{-3}$&	$m$ \\
			\hline
			$d$ &	$223 \times 10^{-3}$ &	$m$ &$m_0$&	$30\times10^{-3}$&	$kg$\\
			\hline
			$I_x$&	$13.2 \times 10^{-3}$&	$N.m.s^2$& $m_1$	&$55\times10^{-3}$	&  $kg$\\
			\hline
			$I_y$	& $12.5 \times 10^{-3}$ &	$N.m.s^2$ &	$m_2$ &	$112\times10^{-3}$& $kg$\\
			\hline
			$I_z$&	$23.5 \times 10^{-3}$&	$N.m.s^2$	& $I_r$	& $33.2 \times 10^{-6}$& $N.m .s^2$\\
			\hline
			$L_0$	& $30\times10^{-3}$ & $m$ &	$L_1$& $70\times10^{-3}$&$m$ \\
			\hline
			$K_{F_1}$ & $1.6\times10^{-5}$ & $kg.m.rad^{-2}$ & $K_{F_2}$&$1.2\times10^{-5}$&$kg.m.rad^{-2}$\\
			\hline
			$K_{F_3}$ &$1.7\times10^{-5}$&$kg.m.rad^{-2}$&$K_{F_4}$& $1.5\times10^{-5}$ &$kg.m.rad^{-2}$\\
			\hline
			$K_{M_1}$ & $3.9\times10^{-7}$ &$kg.m^{2}.rad^{-2}$&$K_{M_2}$ & $2.8\times10^{-7}$ &$kg.m^{2}.rad^{-2}$\\
			\hline
			$K_{M_3}$ &$4.4\times10^{-7}$& $kg.m^{2}.rad^{-2}$ & $K_{M_4}$&$3.1\times10^{-7}$ & $kg.m^{2}.rad^{-2}$ \\
			\hline
		\end{tabular}
	\end{center}
\end{table}

The desired trajectories of the end-effector are generated to follow a circular helix,  while its orientation follows quintic polynomial trajectories \cite{spong2006robot}. Parameters of the proposed algorithm are presented in Table \ref{tab:DOb_par}. The controller is tested to achieve task space trajectory tracking under the effect of the contact force, wind disturbances, and measurement noise. 
\begin{table}[!t]
	\caption{Controller parameters}
	\label{tab:DOb_par}
	\begin{center}
		\setlength{\tabcolsep}{0.01pt}
		\begin{tabular}{|c|c|c|c|}
			\hline
			$Parameter$ & $Value$ & $Par.$ & $Val.$\\ 
			\hline
			$M_{n_\zeta}$	& $diag\{0.02, 0.02, 2, 0.05, 0.01, 0.01\}$ &  	$A_{e_1}$ & $0.16$ \\
			\hline
			$S_{c,d}$ & $diag\{20, 20, 30, 50, 100, 500\}$& $D_{c}$ & $0.01 I_6$ \\
			\hline
			$D_{c,d}$ & $diag\{15, 15, 25, 100, 100, 100\}$& $g_{v_i}$ & $100$ \\
			\hline
			$M_{n_\sigma}$	& $diag\{0.05, 0.05\}$ & $A_{e_2}$ & $0.032$ \\
			\hline
			$\eta^{min}_{h}$ & $0.8$  & $\gamma_{g}$ & $5$ \\
			\hline
			$S_{c}$	& $0.1 I_6$  &  $z_0$& $1$ \\
			\hline
			$K_{p_\sigma}$ / $K_{d_\sigma}$& $20 I_2$ & $V_{w_{z_0}}$ & $3$ \\
			\hline
		\end{tabular}
	\end{center}
\end{table}
\subsection{Estimation of the End-effector Contact Force}
\begin{figure}[!t]
	\centering
	\includegraphics[width=0.95\columnwidth]{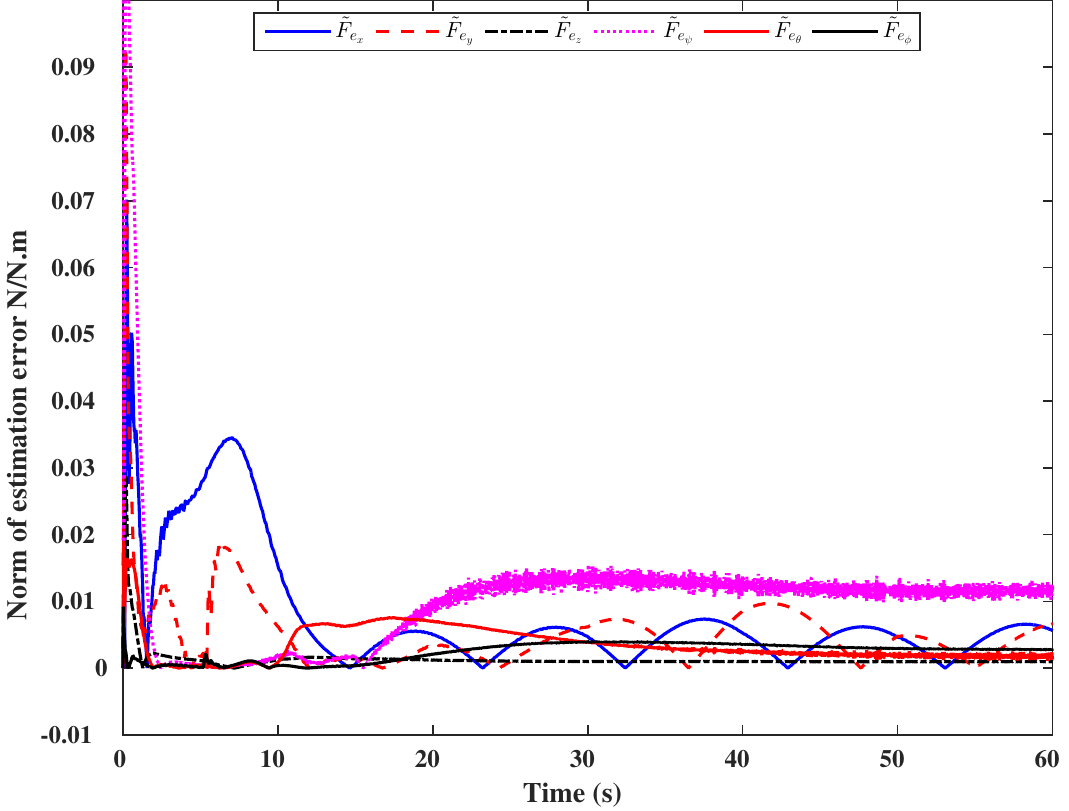}
	\caption{Error norm of estimation of the environment dynamics/contact force}
	\label{fig:Fe}
\end{figure}

Fig. \ref{fig:Fe} shows the response of the proposed algorithm to estimate the environment effect/end-effector contact force. From this figure, it is possible to recognize that the norm of the end-effector generalized force has maximum value of $1$ N/N.m at the beginning of operation due to the time taken by the DOb to estimate the system dynamics and external forces. This initial time is about $3$ s. After that period the error norm decreases gradually. The norm of estimation error in both $x$ and $y$ directions have maximum value of $0.03$ N with sinusoidal shape which is due to the sinusoidal motion in theses axes. While the norm value in $z$ direction have maximum value of $0.005$ N. In both $\phi$ and $\theta$ directions, the norms reach value of $0.007$ N.m. The maximum value of the norm in the $\psi$ direction is about $0.0015$ N.m. Thus, it is possible to appreciate the estimation performance of the end-effector generalized forces. Consequently, one can contend that the second control objective is achieved.

\subsection{Impedance Control}
\begin{figure*}[!t]
	\centering
	\begin{tabular}{c}
		\subfloat{\includegraphics[width=0.7\textwidth]{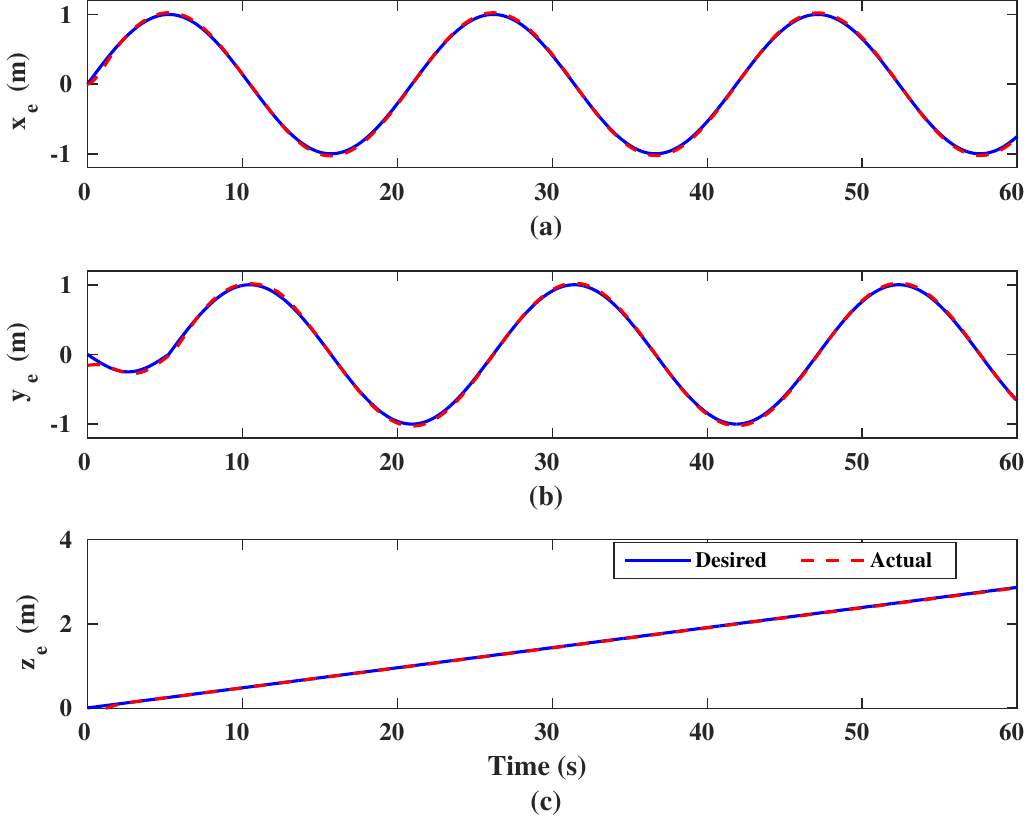}}\\
		\subfloat{\includegraphics [width=0.7\textwidth]{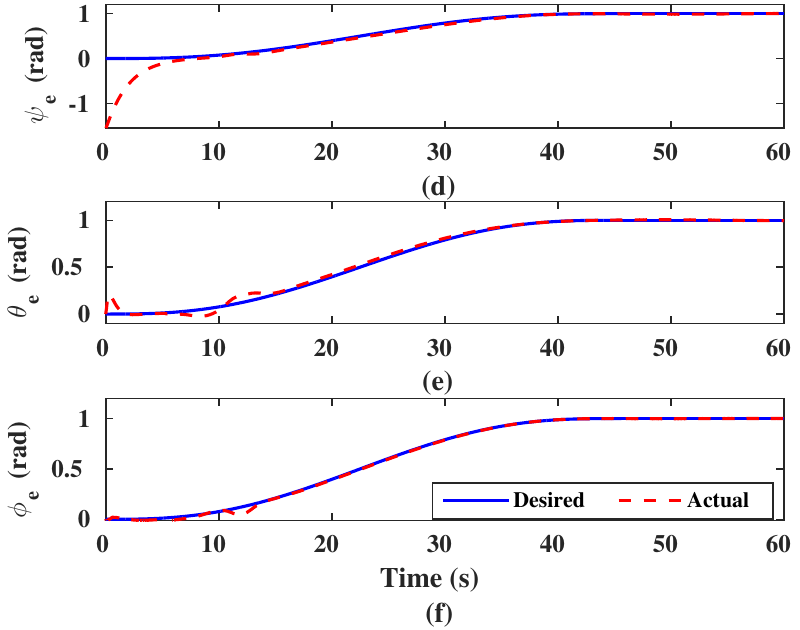}}
	\end{tabular}
	\caption{The actual response of the end-effector position and orientation: 
		a) $x_{e}$, b) $y_{e}$, c) $z_{e}$, d) $\psi_{e}$, e) $\theta_{e}$, and f) $\phi_{e}$}
	\label{fig:inv_kin_pp_results}
\end{figure*}

Fig. \ref{fig:inv_kin_pp_results} shows the response of the system in the task space (the actual end-effector position and orientation can be found from the forward kinematics). From this figure, it is possible to recognize that the controller has good tracking of the desired trajectories of the end-effector (i.e., the tracking error tends to zero as with time). Moreover, it is clear that the capability of the proposed technique to recover the trajectory tracking in the presence of parameters uncertainties which are applied at instant $15$ s. As we see, in the $x$, $y$, $z$, and $\psi$ directions, there is no effect on the tracking. However, in the $\theta$ and $\phi$ directions, the uncertainty effect appears and the controller can recover the tracking quickly. 

\begin{figure}[!t]
	\centering
	\includegraphics[width=0.95\columnwidth]{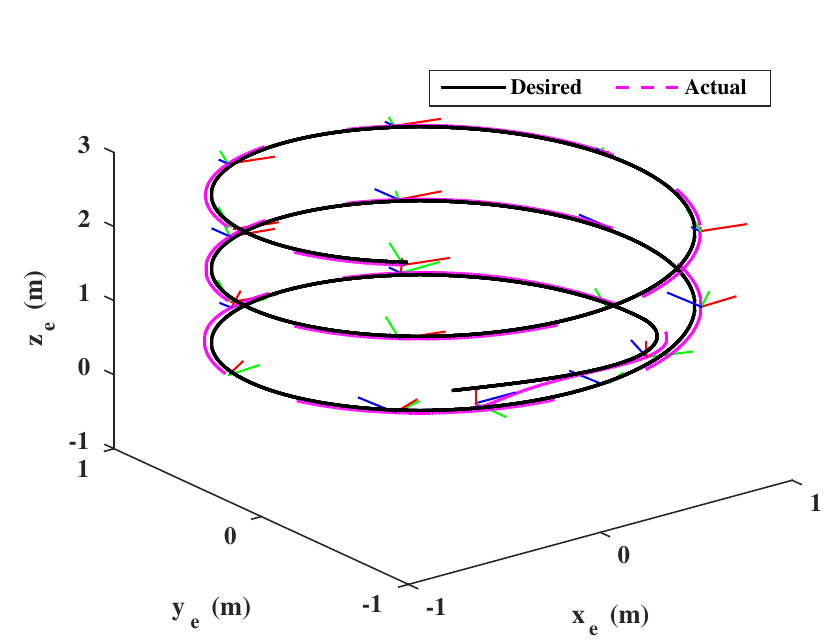}
	\caption{3D trajectory of the end-effector pose (The marker represents the end-effector orientation; Green, Blue, and Red for x-,y, and z-axis, respectively)}
	\label{fig:DOb_3d}
\end{figure}

Fig. \ref{fig:DOb_3d} shows the motion of the end-effector in the 3D dimension (The markers represent the orientation  ). These results show that the proposed impedance motion control scheme provides a robust performance to track the desired end-effector trajectories as well as achieve the desired compliance/impedance effect on the environment taken into consideration the external disturbances and noises. As a result, one can claim that the three control objectives are achieved. 
\section{Conclusion} \label{se:concl}
The problem of the contact force estimation and impedance control of an aerial manipulation robot is presented with a new solution. A brief presented of the system modeling is given. DOb-based system linearization is implemented in the quadrotor/joint space. A DOb is used in the inner loop is to achieve robust linear input/output behavior of the system by compensating disturbances, measurement noise, and uncertainties. Contact force/environment impedance is estimated based on FTRLS and DOb which appear efficient estimation results and stability guarantee. Then, a linear impedance control is designed and implemented in the task space. The inverse kinematics problem is solved by utilization of the system Jacobian. The controller is tested to achieve trajectory tracking under the effect of external wind disturbances, parameters uncertainty, and measurement noise. Numerical results enlighten the efficiency of the proposed control scheme.
\bibliographystyle{asmems4} 
\bibliography{My_Ref}

\end{document}